\documentclass[conference]{IEEEtran}

\usepackage{graphicx}
\usepackage{amsmath}
\usepackage{booktabs}
\usepackage{multirow}
\usepackage{hyperref}

\title{Detection of Hate and Threat in Digital Forensics: A Case-Driven Multimodal Approach}

\author{
\IEEEauthorblockN{Ponkoj Chandra Shill}
\IEEEauthorblockA{
Computer Science and Engineering\\
University of Nevada, Reno\\
Reno, NV, USA\\
}
}

\begin{document}
\maketitle

\begin{abstract}
Digital forensic investigations increasingly rely on heterogeneous evidence such as images, scanned documents, and contextual reports. These artifacts may contain explicit or implicit expressions of harm, hate, threat, violence, or intimidation, yet existing automated approaches often assume clean text input or apply vision models without forensic justification. This paper presents a case-driven multimodal approach for hate and threat detection in forensic analysis. The proposed framework explicitly determines the presence and source of textual evidence, distinguishing between embedded text, associated contextual text, and image-only evidence. Based on the identified evidence configuration, the framework selectively applies text analysis, multimodal fusion, or image-only semantic reasoning using vision–language models with vision transformer backbones (ViT). By conditioning inference on evidence availability, the approach mirrors forensic decision-making, improves evidentiary traceability, and avoids unjustified modality assumptions. Experimental evaluation on forensic-style image evidence demonstrates consistent and interpretable behavior across heterogeneous evidence scenarios.
\end{abstract}


\section{Introduction}

Digital forensic investigations increasingly involve heterogeneous evidence sources, including images, screenshots, scanned documents, and narrative reports. Such artifacts may contain linguistic or visual cues indicating hate speech, threats, harassment, or violent intent, all of which are critical for risk assessment and legal decision-making. In practice, however, automated analysis tools used in forensic workflows often treat evidence modalities in isolation or assume the availability of clean and reliable textual input.

Most prior research on hate and threat detection focuses on well-structured text, such as social media posts or online comments. In contrast, forensic evidence frequently presents more complex data situations. Text may be embedded within images and extracted through noisy optical character recognition (OCR), appear as contextual descriptions external to the visual artifact, or be entirely absent. In other cases, images with no textual content may still convey threatening or violent intent through visual semantics alone. Applying a single analysis strategy across all such scenarios can lead to unreliable conclusions and reduced forensic defensibility.

At the same time, advances in machine learning have enabled strong unimodal models for both text and image analysis. Large pretrained language models such as BERT have significantly improved text-based hate and threat classification \cite{devlin2019bert}, while vision-language models such as CLIP have demonstrated effective zero-shot image classification across broad semantic categories \cite{radford2021learning}. However, applying these models independently to forensic evidence produces fragmented interpretations that fail to capture cross-modal dependencies.


This paper addresses these challenges by proposing a case-driven multimodal forensic analysis pipeline for hate and threat detection. Rather than immediately applying machine learning models, the approach first determines what forms of evidence are present and which analyses are justified. The pipeline explicitly distinguishes between embedded textual evidence within images, associated contextual text from forensic reports or metadata, and image-only evidence. Image content is analyzed using a pretrained vision–language model, while embedded and associated text are analyzed using a zero-shot text classification model.

The central challenge in forensic multimodal analysis is not model expressiveness, but evidentiary legitimacy. Semantic inference must be conditioned on evidence availability to avoid unsupported assumptions that are unacceptable in forensic settings. Accordingly, modality-specific analyses are selectively applied and reconciled through an explicit and auditable decision process that mirrors forensic reasoning.

To ensure that multimodal evidence can be combined transparently, modality-specific confidence scores are aggregated using a simple weighted scheme, illustrated as follows:
\begin{equation}
s^{(\mathrm{fused})}_{\ell_k} =
\frac{\sum_{m} w_m s^{(m)}_{\ell_k}}{\sum_{m} w_m}
\end{equation}
This formulation is shown here to illustrate the explicit decision logic of the framework. All variables, modality conditions, and fusion rules are formally defined in Section~III.

\textbf{Contributions.} This research makes the following contributions:
\begin{itemize}
    \item A case-driven forensic framework that explicitly distinguishes embedded text, associated contextual text, and image-only evidence prior to semantic analysis.
    \item A modality-aware routing and score-level fusion strategy grounded in forensic reliability rather than end-to-end learned fusion.
    \item An experimental evaluation demonstrating consistent performance across heterogeneous forensic evidence configurations without task-specific training.
\end{itemize}

The remainder of this paper is organized as follows. Section~II reviews prior work on unimodal and multimodal hate detection, vision-language models, and forensic text analysis. Section~III presents the proposed case-driven multimodal forensic analysis framework and its underlying decision logic. Section~IV describes the application workflow and experimental setup. Section~V reports experimental findings and analyzes the impact of multimodal fusion. Section~VI discusses implications and limitations, and Section~VII concludes the paper.

\section{Related Work}
\label{sec:related}

This section reviews prior research relevant to multimodal hate and threat detection from three perspectives: (i) unimodal and multimodal hate speech detection, (ii) vision--language and zero-shot approaches, and (iii) forensic-oriented text analysis and evidence triage. The discussion highlights how existing work motivates multimodal reasoning while exposing gaps addressed by the proposed case-driven forensic pipeline.

\subsection{Unimodal and Multimodal Hate Detection}

Early research on hate speech and abusive language detection primarily focused on text-only data, leveraging traditional machine learning methods and, more recently, large pretrained language models such as BERT \cite{devlin2019bert}. These approaches assume access to clean, well-structured text, an assumption that often does not hold in forensic investigations where text may be fragmented, noisy, or entirely absent.

The limitations of unimodal approaches motivated a shift toward multimodal hate speech detection. A seminal contribution in this area is the Hateful Memes benchmark introduced by Kiela \emph{et al.} \cite{kiela2020hateful}, which demonstrated that hateful intent may emerge only through the interaction of image and text, even when each modality appears benign in isolation. 

Several studies have explored multimodal fusion strategies for hate speech detection. Gomez \emph{et al.} \cite{gomez2020multimodal} investigated early fusion of image and text embeddings, while Boishakhi \emph{et al.} \cite{boishakhi2021multi} demonstrated that integrating heterogeneous modalities provides complementary signals for detecting hateful content. More broadly, surveys of multimodal machine learning highlight the potential of joint reasoning across modalities \cite{baltrusaitis2018multimodal}. However, these approaches typically assume fixed input structures, rely on end-to-end learned fusion, and offer limited interpretability, posing challenges for forensic settings where modality availability, uncertainty, and evidentiary traceability must be explicitly managed.

\subsection{Vision-Language Models and Zero-Shot Multimodal Analysis}

Recent advances in pretrained vision--language models have enabled flexible multimodal analysis without task-specific retraining. CLIP \cite{radford2021learning} introduced contrastive pretraining between images and natural language, enabling zero-shot image classification via prompt-based inference. This capability has been adopted for multimodal hate detection. Arya \emph{et al.} \cite{arya2024clip} demonstrated that CLIP-based models with prompt engineering can effectively identify hateful content in memes, highlighting the suitability of vision--language models for semantic image analysis.

Beyond CLIP, zero-shot multimodal frameworks aim to generalize to unseen targets and evolving forms of hate. Zhu \emph{et al.} proposed a multimodal zero-shot hateful meme detection framework that enhances adaptability to novel targets \cite{zhu2024zeroshot}. Yamagishi \cite{yamagishi2024simple} further showed that simpler prompt formulations can outperform complex designs in zero-shot hate detection, emphasizing robustness and generalization. Despite their effectiveness, these approaches are typically evaluated on benchmark datasets with fixed data configurations and do not explicitly model the heterogeneous evidence scenarios common in forensic investigations.

\subsection{Forensic Text Analysis and Evidence-Centric Triage}

In forensic workflows, textual evidence typically appears as unstructured narrative reports rather than clean social media posts. Prior work has shown that such data can be effectively analyzed using NLP techniques. Kučerová \emph{et al.} \cite{kucerova2020crime} demonstrated crime report classification using machine learning, while Karystianis \emph{et al.} \cite{karystianis2018ie} focused on automated extraction of entities such as weapons, locations, and actions from police reports. These studies motivate treating associated forensic text as a distinct and semantically rich evidence source.

Automated evidence triage has also been emphasized in digital forensics to manage large volumes of data prior to human review. Du \emph{et al.} \cite{du2020forensics} surveyed deep learning approaches for prioritizing and categorizing evidentiary artifacts. However, most existing systems focus on coarse file-based or content-based sorting, whereas the present work performs semantic triage conditioned on evidence availability and modality configuration.

Text embedded within images presents additional challenges due to OCR noise, which can significantly degrade downstream NLP performance. Treating OCR-derived text as equivalent to clean input can therefore lead to unreliable inference. By explicitly separating OCR-derived embedded text from associated contextual text and visual evidence, the proposed framework aligns automated analysis with forensic reasoning and supports transparent, auditable decision-making.

Overall, existing multimodal hate detection and forensic analysis methods do not explicitly model heterogeneous evidence configurations or provide transparent fusion mechanisms. The proposed case-driven pipeline addresses this gap by routing evidence based on modality availability and fusing modality-specific outputs within a shared frozen label space.

\section{Methodology}
\label{sec:methodology}

\begin{figure*}[t]
    \centering
    \includegraphics[width=0.9\textwidth]{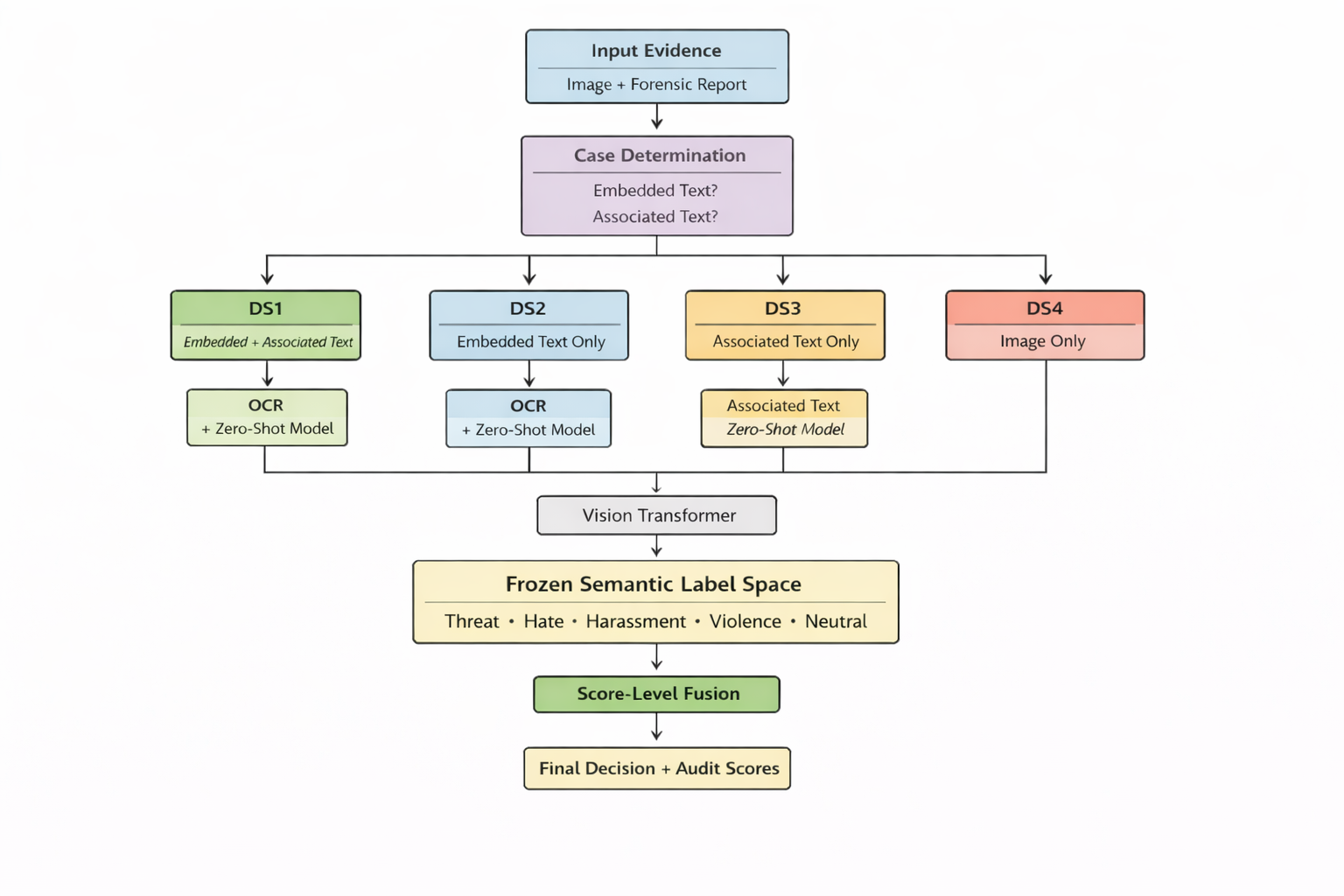}
    \caption{Case-driven multimodal forensic analysis pipeline. Image evidence and forensic reports are first evaluated to determine the presence of embedded and/or associated text, resulting in one of four data situations (DS1--DS4). Depending on the case, OCR and zero-shot text models are applied alongside vision-based analysis. All modality-specific outputs are mapped into a shared frozen semantic label space and combined through score-level fusion to produce a final decision with auditable scores.}
    \label{fig:methodology}
\end{figure*}

This section presents the proposed case-driven multimodal forensic analysis framework. The methodology is designed to operate on heterogeneous forensic evidence while preserving interpretability, auditability, and robustness to missing modalities. An overview of the proposed case-driven multimodal pipeline is shown in Fig.~\ref{fig:methodology}.

\subsection{Problem Definition}
\label{sec:problem}

The objective of this work is to perform semantic assessment of image-centric forensic evidence for hate, threat, harassment, and violence-related categories. Each forensic evidence item is associated with a unique media identifier and consists of an image \(I\), along with zero or more textual components. Textual components may include embedded text extracted from the image using optical character recognition (OCR), denoted as \(T^{\text{ocr}}\), and associated contextual text obtained from forensic reports, metadata, or investigator notes, denoted as \(T^{\text{assoc}}\).

Formally, an evidence item is represented as
\[
E = (I, T^{\text{ocr}}, T^{\text{assoc}}),
\]
where one or both textual components may be absent. Given \(E\), the task is to estimate the likelihood that the evidence corresponds to predefined forensic-relevant semantic categories while satisfying constraints of interpretability, traceability, and robustness to missing data.

Rather than producing a single opaque prediction, the system generates modality-specific confidence scores and derives a final decision through an explicit fusion process. The approach avoids task-specific retraining and operates in a zero-shot setting to ensure adaptability to evolving forensic datasets and threat categories.

\subsection{Forensic Evidence Scenarios}
\label{sec:cases}

Forensic evidence exhibits substantial variability in modality availability and quality. To reflect this reality, the proposed framework explicitly distinguishes between three forensic evidence scenarios, referred to as cases.

\textbf{Case 1: Image with Embedded Text.} In this scenario, the image $I$ contains text that is visually embedded within the artifact, such as screenshots of messages, scanned documents, or edited images. Optical Character Recognition (OCR) is applied to extract this content by identifying text regions and mapping visual glyphs to machine-readable symbols, producing OCR-derived text $T_{\text{ocr}}$. In forensic settings, OCR output is often affected by noise arising from low resolution, compression artifacts, non-standard fonts, occlusion, or intentional manipulation, leading to recognition errors and incomplete text. Despite these limitations, embedded text frequently conveys critical semantic information related to threats, harassment, or intent that may not be evident from visual semantics alone. Accordingly, the proposed framework analyzes both the image $I$ and the OCR-derived text $T_{\text{ocr}}$ independently, treating OCR text as a distinct modality rather than merging it with cleaner contextual text. This design preserves uncertainty introduced during text extraction and enables explicit, evidence-aware fusion while reflecting common forensic scenarios involving screenshots and scanned materials. An example of this scenario is shown in Fig.~2, where threatening intent is conveyed through visually embedded text rather than visual objects or scenes.

\begin{figure}[t]
    \centering
    \includegraphics[width=0.9\linewidth,,height=4cm,keepaspectratio]{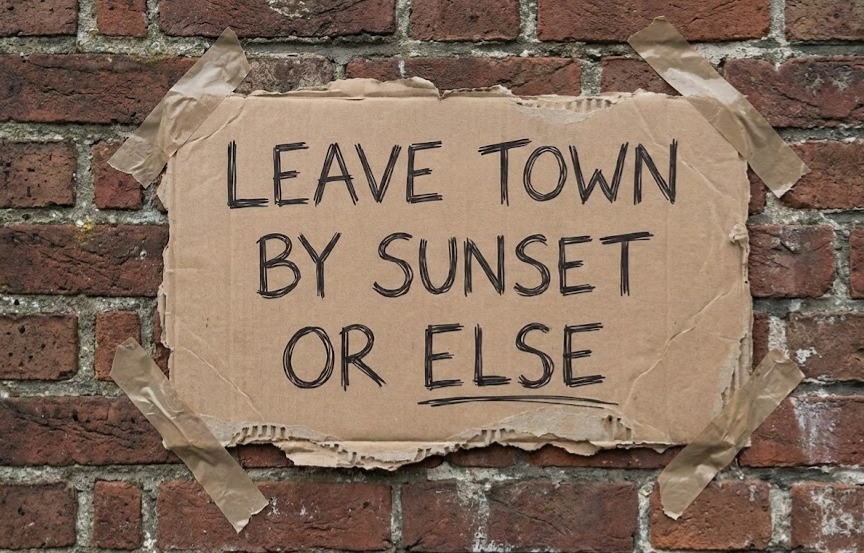}
    \caption{Example of Case 1 (Image with Embedded Text). Threatening intent is conveyed through visually embedded text extracted via OCR.}
    \label{fig:case1_example}
\end{figure}

\textbf{Case 2: Image with Associated Text.} In this scenario, the image $I$ is accompanied by contextual text $T_{\text{assoc}}$ that is not visually embedded in the image but is semantically linked through forensic reports or metadata. Associated text is obtained from narrative descriptions, chat transcripts, or suspect or victim statements that reference specific media items. The association is established deterministically using shared media identifiers, filenames, or evidence IDs recorded during evidence collection. To ensure contextual relevance, the image and text must originate from the same communication thread or report segment, and their timestamps must fall within a fixed temporal window of 120 seconds, preventing cross-context contamination in accordance with common forensic practice. Because associated text is human-authored rather than OCR-derived, it is typically more complete and linguistically coherent. Accordingly, $T_{\text{assoc}}$ is treated as a higher-confidence textual modality and analyzed independently from OCR-derived text, preserving clear provenance and enabling transparent, evidence-aware fusion. An example of this scenario is shown in Fig.~3, where forensic relevance emerges from associated textual context despite the absence of embedded text in the image.

\begin{figure}[t]
    \centering
    \includegraphics[width=0.9\linewidth,,height=4cm,keepaspectratio]{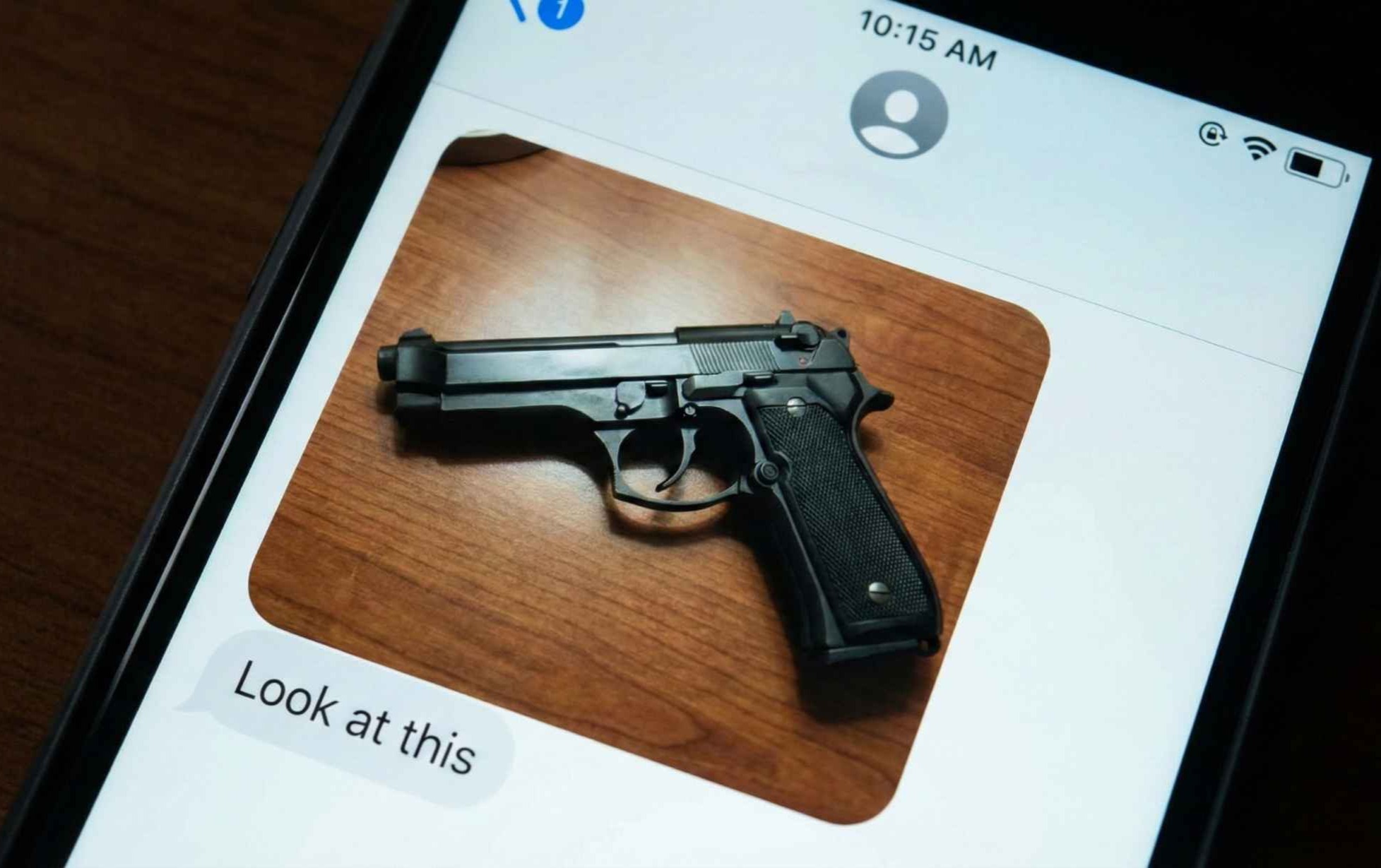}
    \caption{Example of Case 2 (Image with Associated Text). The image contains no embedded text; semantic interpretation relies on associated contextual text and metadata within the same communication thread.}
    \label{fig:case2_example}
\end{figure}

\textbf{Case 3: Image-Only Evidence.}  
In this scenario, the forensic evidence consists solely of an image \(I\) with no usable textual information available, meaning that neither embedded text suitable for optical character recognition (OCR) nor associated contextual text from forensic reports or metadata is present. This situation commonly arises with standalone photographs or media items extracted without accompanying narratives. Under this condition, semantic assessment relies exclusively on visual content, which is analyzed using a pretrained vision--language model to infer forensic-relevant categories based on visual cues such as objects, scenes, or symbolic elements. No text-based inference or multimodal fusion is performed, and the framework does not attempt to infer missing context, thereby preserving forensic defensibility by restricting analysis to observable evidence only.

These case definitions guide modality routing and determine which analyses are performed for each evidence item. Importantly, multimodal fusion is applied only when justified by evidence availability, rather than assuming a fixed multimodal structure for all inputs.
\subsection{Modality: Derived Data Situations}
\label{sec:datasituations}

\begin{figure}[t]
    \centering
    \includegraphics[width=\linewidth]{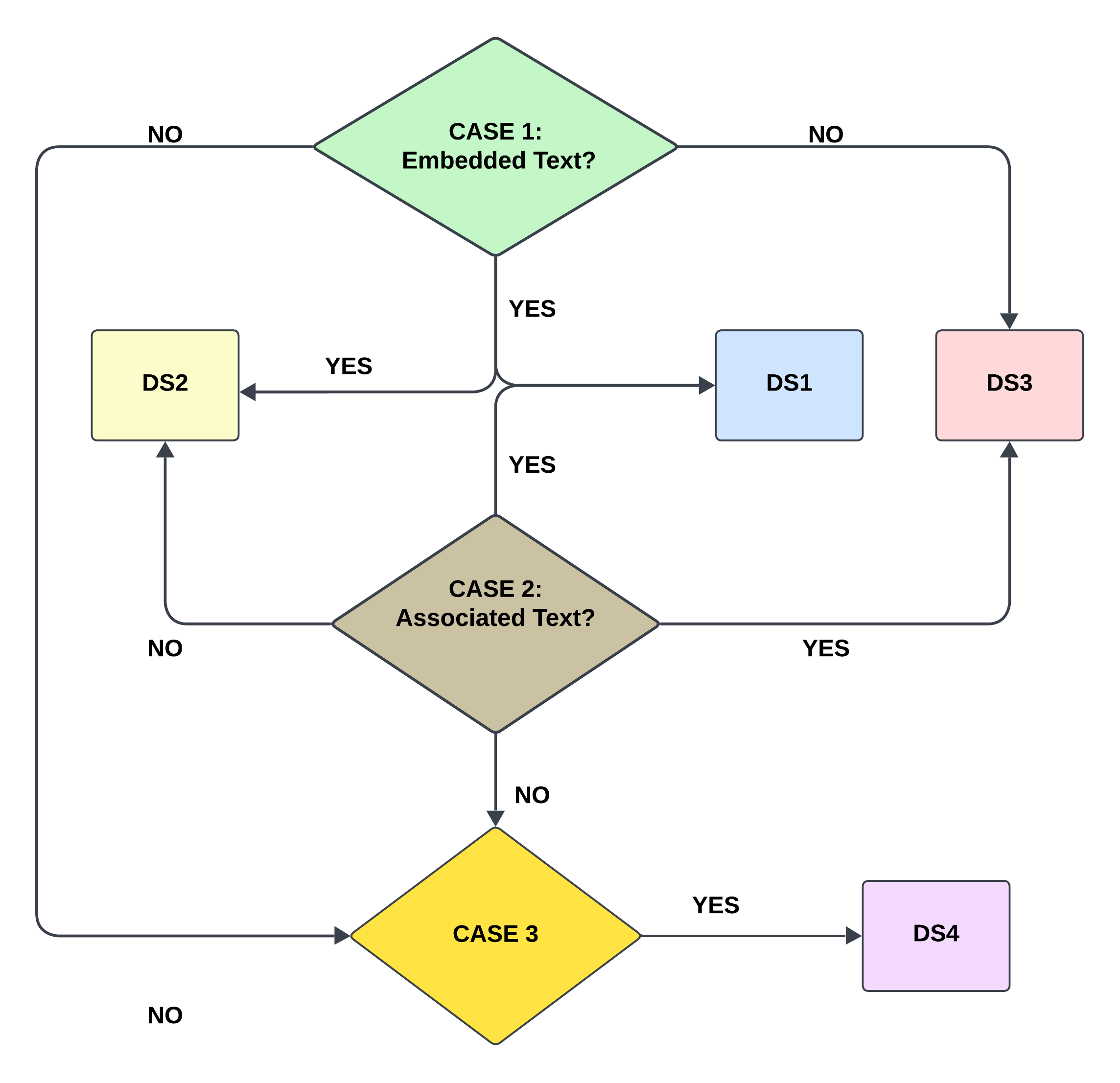}
    \caption{Case-based modality determination logic. The framework first evaluates the presence of embedded text within the image, followed by the availability of associated contextual text from forensic reports or communication threads. Based on these decisions, each image is routed to one of four derived data situations (DS1--DS4), ensuring that subsequent analysis is conditioned on legitimately available evidence.}
    \label{fig:case_determination}
\end{figure}

Based on the identified forensic evidence case, the framework derives a corresponding data situation that determines which modality-specific analyses are executed. These data situations formalize the routing logic between evidence availability and model execution. The case/modality determination logic used to route image evidence based on the availability of embedded and associated text is illustrated in Fig.~\ref{fig:case_determination}.

\textbf{DS1: Embedded and Associated Text Present.}  
When both OCR-derived embedded text \(T^{\text{ocr}}\) and associated contextual text \(T^{\text{assoc}}\) are available, multimodal analysis is performed using image semantics, OCR text, and associated text.

\textbf{DS2: Embedded Text Only.}  
When only OCR-derived embedded text is available, multimodal analysis combines visual semantics with OCR-based text inference.

\textbf{DS3: Associated Text Only.}  
When only associated contextual text is available, multimodal analysis combines image semantics with associated text inference.

\textbf{DS4: No Text Available.}  
When no textual information is available, semantic assessment is performed using image-only analysis via a vision--language model.

This explicit separation between forensic evidence cases and derived data situations ensures that model execution remains traceable, interpretable, and aligned with forensic standards.

\subsection{Frozen Label Space}
\label{sec:frozen}

All analyses in the proposed framework map evidence to a shared \emph{frozen semantic label space}
\[
\mathcal{L} = \{\ell_1, \ell_2, \dots, \ell_K\},
\]
where each label \(\ell_k\) corresponds to a forensic-relevant semantic category. In this study, the frozen label space consists of the following categories:
\begin{itemize}
    \item \textit{Threat of Violence}
    \item \textit{Harassment or Intimidation}
    \item \textit{Hate or Bias-Based Content}
    \item \textit{Incitement or Endorsement of Violence}
    \item \textit{Weapon-Related Content}
    \item \textit{Criminal Admission or Description}
    \item \textit{Sexual Violence or Exploitation}
    \item \textit{Abusive or Obscene Language}
    \item \textit{Self-Harm or Suicide Risk}
    \item \textit{Neutral or Contextual Content}
\end{itemize}

The label space is defined prior to analysis and remains unchanged throughout system execution. In this work, the frozen label space represents a fixed semantic decision layer. All modality-specific models operate in a zero-shot inference setting and project their outputs into this shared label space.

For image analysis, each label is represented as a fixed natural-language prompt and evaluated using a pretrained vision transformer (ViT) language model based on Contrastive Language Image Pretraining (CLIP). Specifically, we use the \texttt{ViT-L/14} architecture pretrained on the \texttt{LAION-2B} dataset (\texttt{laion2b\_s32b\_b82k}). Image embeddings are compared against prompt embeddings to produce label-specific confidence scores.

For text analysis, the same frozen labels are used as candidate hypotheses in a zero-shot text classification model based on a pretrained transformer architecture (e.g., DeBERTa-v3-large zero-shot models). This model computes label relevance scores for both OCR-derived text and associated contextual text using identical semantic definitions.

By fixing the semantic labels and their textual descriptions, the framework enforces semantic consistency across image-based and text-based assessments and prevents label drift across modalities and models. This design enables direct comparison and score-level fusion of modality-specific outputs while preserving interpretability, auditability, and forensic defensibility.

\subsection{Modality-Specific Analysis}
\label{sec:modalities}

Each available modality is analyzed independently using pretrained zero-shot models.

\subsubsection{Image Analysis}

Visual content is analyzed using a pretrained vision transformer (ViT)–based vision–language model based on Contrastive Language Image Pretraining (CLIP). Given an image \(I\) and a set of natural language prompts derived from the frozen label space \(\mathcal{L}\), the model computes similarity scores between the image embedding and each label prompt embedding. This produces an image-based score vector
\[
\mathbf{s}^{\text{img}} =
\left[
s^{\text{img}}_{\ell_1}, s^{\text{img}}_{\ell_2}, \dots, s^{\text{img}}_{\ell_K}
\right],
\]
where \(s^{\text{img}}_{\ell_k}\) represents the confidence assigned to label \(\ell_k\) based on visual semantics.

\subsubsection{Text Analysis}

Textual content, including both OCR-derived text \(T^{\text{ocr}}\) and associated contextual text \(T^{\text{assoc}}\), is analyzed using a zero-shot text classification model based on a pretrained transformer architecture (e.g., DeBERTa-v3-large zero-shot models). Each text segment is evaluated independently against the frozen label space \(\mathcal{L}\), producing a score vector
\[
\mathbf{s}^{\text{text}} =
\left[
s^{\text{text}}_{\ell_1}, s^{\text{text}}_{\ell_2}, \dots, s^{\text{text}}_{\ell_K}
\right].
\]

OCR-derived text and associated text are treated as distinct inputs due to differences in noise characteristics and semantic reliability. When multiple text segments are available, scores are aggregated at the media identifier level.

\subsection{Score-Level Fusion}
\label{sec:fusion}

When multiple modalities are available for a given evidence item, modality-specific confidence scores are combined using explicit score-level fusion. For each label \(\ell_k \in \mathcal{L}\), the fused score is computed as
\[
s^{\text{fused}}_{\ell_k} =
\frac{\sum_{m \in \mathcal{M}} w_m \, s^{(m)}_{\ell_k}}
{\sum_{m \in \mathcal{M}} w_m},
\]
where \(\mathcal{M} \subseteq \{\text{img}, \text{ocr}, \text{assoc}\}\) denotes the set of available modalities, \(s^{(m)}_{\ell_k}\) is the confidence score produced by modality \(m\), and \(w_m\) is a fixed modality-specific weight.

In this study, fusion weights are fixed and selected based on the relative forensic reliability of each modality rather than learned from data. Specifically, we assign \(w_{\text{img}} = 1.0\) to image-based inference, \(w_{\text{ocr}} = 1.0\) to OCR-derived embedded text, and \(w_{\text{assoc}} = 1.2\) to associated contextual text extracted from forensic reports. The higher weight for associated text reflects its typically greater semantic clarity and contextual completeness compared to OCR-derived text.

Modalities that are not available for a given evidence item are excluded from the fusion process rather than implicitly penalized. That is, if a modality \(m\) is absent, it is not included in \(\mathcal{M}\), and its contribution is effectively zero. This design ensures that missing modalities do not bias the fused scores and that fusion behavior adapts transparently to different forensic evidence scenarios.

Score-level fusion is performed independently for each label in the frozen label space, and both modality-specific scores and fused scores are retained to support auditability, reproducibility, and expert review.

\subsection{Decision Logic and Output Representation}
\label{sec:output}

The final semantic assessment for an evidence item is determined by selecting the label with the maximum fused score,
\[
\ell^{*} = \arg\max_{\ell_k \in \mathcal{L}} s^{\text{fused}}_{\ell_k}.
\]

In addition to the final label, the framework outputs complete modality-specific score vectors, fused scores, and metadata associated with the media identifier. Results are stored in structured tabular formats to support auditability, reproducibility, and downstream forensic analysis.

By explicitly modeling forensic evidence scenarios, preserving modality-level outputs, and employing transparent score-level fusion, the proposed methodology aligns automated multimodal analysis with the practical requirements of digital forensic investigations.



\section{Application and Experimental Findings}

\subsection{Application Workflow}
The proposed framework is applied to forensic evidence through a structured, case-driven workflow. Each media item is first ingested with its corresponding metadata and assigned a unique media identifier. Based on the presence or absence of embedded text and associated contextual text, the system determines the applicable forensic case and derives the corresponding data situation. Image evidence is analyzed using a pretrained vision--language model, while textual evidence is analyzed using zero-shot text classification. All modality-specific outputs are mapped to the frozen semantic label space and stored in structured CSV artifacts, enabling traceability, reproducibility, and post hoc review.

The system produces three primary outputs: (i) image-based semantic scores generated by the vision--language model, (ii) text-based semantic scores generated from OCR-derived or associated textual evidence, and (iii) fused scores combining available modalities. Each output preserves per-label confidence values and explicitly records which modalities contributed to the final decision.

\subsection{Experimental Setup}
Experiments were conducted using heterogeneous forensic-style evidence consisting of images, OCR-extracted text, and associated textual descriptions derived from forensic reports or communication logs. No task-specific training or fine-tuning was performed. Image analysis was conducted using a CLIP-based vision--language model with a ViT-L/14 backbone pretrained on the LAION-2B dataset. Text analysis was performed using a transformer-based zero-shot classification model (DeBERTa-v3-large-based), applied consistently to both OCR-derived text and associated contextual text.

All experiments were executed in inference mode on commodity hardware with GPU acceleration when available. Fusion weights were fixed a priori, and missing modalities were excluded from fusion in accordance with the derived data situation. This setup reflects realistic forensic deployment conditions where labeled training data may be unavailable and interpretability is critical.

\subsection{Case-Based Findings}
Across evaluated evidence items, the system exhibited distinct behavior aligned with the defined forensic cases. In Case~1 scenarios, where threatening or abusive language was embedded within images, OCR-derived text provided essential semantic cues that were not reliably inferred from visual content alone. These cases demonstrated the importance of explicitly modeling OCR uncertainty while still allowing embedded text to influence final assessments.

In Case~2 scenarios, image-only semantics were often ambiguous, with visual content such as objects or scenes requiring contextual interpretation. Associated textual evidence extracted from forensic reports or communication threads frequently clarified intent, shifting semantic interpretation toward threat, intimidation, or criminal relevance. The use of deterministic metadata mapping and temporal constraints ensured that only contextually valid text influenced analysis.

In Case~3 scenarios, where no textual evidence was available, the framework relied solely on visual semantics. Image-only analysis successfully identified visually salient forensic categories such as weapon-related content, while appropriately limiting inference in the absence of linguistic cues. The system did not attempt to infer missing context, preserving conservative and defensible behavior.

\subsection{Multimodal Fusion Outcomes}
Score-level fusion enabled the framework to reconcile complementary and occasionally conflicting signals across modalities. In scenarios where both image and textual evidence were available, fused scores often exhibited increased confidence in forensic-relevant categories compared to unimodal analysis. When modalities disagreed, per-modality scores revealed the source of divergence, allowing investigators to inspect whether ambiguity arose from OCR noise, contextual language, or visual interpretation.

Importantly, fusion behavior adapted transparently to evidence availability. When one or more modalities were absent, fusion reduced to the remaining inputs without penalization. Retaining both fused and modality-specific scores ensured that automated decisions remained explainable and auditable.

\section{Results}

\subsection{Pipeline Agreement with Manual Annotation}
\label{sec:result1}

The correctness of the proposed framework was evaluated using manually annotated ground truth. Each image evidence item was reviewed together with all available textual context, including OCR-derived embedded text and associated contextual text when present. Based on this review, each item was assigned a binary ground-truth label indicating whether it contained \emph{Harmful Content} (hate, threat, harassment, or violent intent) or \emph{Non-Harmful Content}. Pipeline predictions were derived from fused semantic scores within the frozen label space and converted into a binary decision consistent with forensic triage use cases.

Performance was evaluated separately for each derived data situation (DS1--DS4) to assess the impact of evidence availability on detection accuracy. Table~\ref{tab:ds_accuracy} summarizes the results.

\begin{table}[t]
\centering
\caption{Pipeline Accuracy Across Derived Data Situations}
\label{tab:ds_accuracy}
\begin{tabular}{lccc}
\hline
\textbf{Modality} & \textbf{Evidence} & \textbf{Correct HT} & \textbf{Accuracy (\%)} \\
\hline
DS1: Embedded + Assoc. Text & 22 & 21 / 22 & 95.45 \\
DS2: Embedded Text Only    & 35 & 33 / 35 & 94.29 \\
DS3: Assoc. Text Only      & 70 & 69 / 70 & 98.50 \\
DS4: Image Only            & 55 & 53 / 55 & 96.36 \\
\hline
\end{tabular}
\end{table}

In \textbf{DS1}, where both embedded text and associated contextual text were available, the pipeline correctly identified 21 out of 22 hate or threat-related images, yielding an accuracy of 95.45\%. The single false negative highlights the residual uncertainty introduced by OCR noise and semantic ambiguity despite rich multimodal evidence.

In \textbf{DS2}, which consists of images containing embedded text only, the pipeline correctly flagged 33 out of 35 images as harmful, achieving an accuracy of 94.29\%. The two false negatives were attributable to incomplete or ambiguous OCR extraction, underscoring the challenges posed by noisy visual text.

In \textbf{DS3}, where images were accompanied by associated contextual text from forensic reports or communication threads, the pipeline achieved near-perfect performance, correctly identifying all 69 images as hate or threat related. This result demonstrates the strong contribution of associated text when combined with visual analysis and validates the higher fusion weight assigned to this modality.

In \textbf{DS4}, which represents image-only evidence with no available textual information, the pipeline correctly identified 53 out of 55 images as harmful, yielding an accuracy of 96.36\%. The remaining false negatives reflect conservative behavior in the absence of linguistic cues, which is appropriate for forensic triage scenarios.

Overall, the results demonstrate that the proposed case-driven multimodal pipeline achieves consistently high accuracy across heterogeneous forensic evidence configurations. Performance improves systematically as richer textual context becomes available, empirically validating the design choices of explicit case modeling, modality-aware routing, and score-level fusion within a frozen semantic label space.

\subsection{Result 2: Impact of Multimodal Fusion}
\label{sec:result2}

This result examines how score-level multimodal fusion alters pipeline decisions relative to image-only inference. Rather than measuring correctness against ground truth, this analysis focuses on \emph{decision dynamics}, specifically how the inclusion of textual evidence (OCR-derived or associated text) influences the final semantic assessment produced by the pipeline.

For each image evidence item, an initial prediction was obtained using image-only inference via the vision--language model. When textual evidence was available, fused predictions were subsequently computed according to the corresponding derived data situation. A decision change was recorded when the fused prediction differed from the image-only prediction in terms of the binary harmful versus non-harmful classification.

Table~\ref{tab:fusion_impact} summarizes the impact of multimodal fusion across data situations.

\begin{table}[t]
\centering
\caption{Impact of Multimodal Fusion on Pipeline Decisions}
\label{tab:fusion_impact}
\begin{tabular}{lccc}
\hline
\textbf{Modality} & \textbf{Evidence} & \textbf{Changed} & \textbf{Rate (\%)} \\
\hline
DS1: Embedded + Assoc. Text & 22 & 5  & 22.73 \\
DS2: Embedded Text Only    & 35 & 12 & 34.29 \\
DS3: Assoc. Text Only      & 70 & 9  & 12.86 \\
DS4: Image Only            & 55 & 0  & 0.00 \\
\hline
\end{tabular}
\end{table}

In \textbf{DS4}, where no textual information was available, multimodal fusion was not applied and all decisions relied exclusively on visual semantics. As expected, no decision changes were observed in this scenario.

In contrast, \textbf{DS1}, \textbf{DS2}, and \textbf{DS3} exhibited measurable decision changes following fusion. The highest change rate was observed in DS2 (34.29\%), where OCR-derived embedded text frequently corrected or refined ambiguous image-only predictions. DS1 showed a change rate of 22.73\%, reflecting the complementary influence of embedded and associated textual cues. DS3 exhibited a lower but non-negligible change rate of 12.86\%, indicating that associated contextual text primarily reinforced visual predictions while occasionally resolving uncertainty.

Importantly, multimodal fusion did not introduce instability or oscillatory behavior. All decision changes were directly traceable to modality-specific score contributions within the frozen semantic label space, enabling transparent inspection of why a decision was altered. This confirms that fusion functions as an evidence-aware refinement mechanism rather than an opaque override of visual inference.

Overall, these findings demonstrate that multimodal fusion meaningfully affects pipeline outcomes when textual evidence is available, while degrading gracefully to image-only inference when it is not. This behavior empirically validates the case-driven routing strategy and score-level fusion design adopted in the proposed framework.

\subsection{Summary of Observations}
The experimental results indicate that a case-driven multimodal framework aligns naturally with forensic reasoning and the heterogeneous nature of digital evidence. By explicitly distinguishing evidence configurations, preserving modality-specific outputs, and enforcing semantic consistency through a frozen label space, the proposed approach enables interpretable and adaptable semantic assessment without reliance on task-specific model training. Importantly, the framework demonstrates robust behavior across varying evidence conditions, degrading gracefully when textual information is unavailable while benefiting from multimodal fusion when context is present. These properties support transparent analysis, reproducibility, and practical deployment in real-world digital forensic investigations.

\section{Discussion}

The primary strength of the proposed framework lies in its explicit case-driven decision logic and close alignment with forensic reasoning. By distinguishing between image-only evidence, images with embedded text, and images accompanied by associated contextual text, the pipeline avoids assumptions common in multimodal systems that treat all inputs as uniformly available or equally reliable. This explicit modeling of evidence configurations enables the system to adapt its analysis strategy based on legitimately available information, improving both interpretability and forensic defensibility.

The experimental results show that multimodal fusion meaningfully influences pipeline decisions when textual evidence is present, while degrading gracefully to image-only inference when it is not. Associated contextual text provides strong semantic cues for intent recognition, whereas OCR-derived text introduces additional uncertainty due to recognition errors. These findings validate the decision to treat OCR-derived text and associated text as distinct modalities with different reliability characteristics and to preserve modality-specific scores alongside fused outcomes.

From a forensic perspective, the framework is best viewed as a structured triage and decision-support tool rather than a replacement for human judgment. By retaining modality-specific outputs and conditioning inference on evidence availability, the system supports analyst oversight, post hoc review, and transparent justification of automated assessments, aligning with established forensic workflows.

Several limitations should be acknowledged. First, the quality of OCR-derived text directly affects performance in cases involving embedded text, and severe recognition errors may lead to missed detections. Second, while vision–language models such as CLIP enable powerful zero-shot semantic reasoning, their internal representations remain partially opaque, limiting fine-grained explainability. Finally, the evaluation relies on manually annotated data and forensic report-derived context, which may introduce subjective bias despite careful review. Addressing these limitations through improved OCR robustness, enhanced explainability mechanisms, and larger-scale validation remains an important direction for future work.

\section{Conclusion}

This research presented a case-driven multimodal pipeline for hate and threat detection in forensic reports and digital evidence. By explicitly determining the availability and source of textual information and routing evidence accordingly, the proposed framework aligns automated semantic analysis with established forensic reasoning and evidentiary practices. The use of a frozen semantic label space and explicit score-level fusion enables transparent, interpretable, and adaptable analysis without reliance on task-specific model training.

Experimental results demonstrate that the pipeline achieves high accuracy when evaluated against manual annotation and that multimodal fusion meaningfully refines decisions when textual context is available. Importantly, the framework maintains conservative and defensible behavior in the absence of text, supporting practical deployment in real-world forensic settings. Future work will explore larger-scale evaluations, tighter integration with forensic reporting and evidence management systems, and extensions to additional forms of multimodal digital evidence. More broadly, this work highlights the importance of evidence-aware system design when deploying machine learning in high-stakes forensic domains.

\section*{Code Availability}
An open-source implementation of the proposed framework is available at \url{https://github.com/CS-Ponkoj/Hate-and-Threat-Detection-in-Forensics}.

\bibliographystyle{IEEEtran}
\bibliography{bib}

@inproceedings{boishakhi2021multi,
  title={Multi-modal Hate Speech Detection using Machine Learning},
  author={Boishakhi, Fariha Tahosin and Shill, Ponkoj Chandra and Alam, Md Golam Rabiul},
  booktitle={2021 IEEE International Conference on Big Data (BigData)},
  year={2021},
  doi={10.1109/BigData52589.2021.9671955}
}

@inproceedings{devlin2019bert,
  title={BERT: Pre-training of Deep Bidirectional Transformers for Language Understanding},
  author={Devlin, Jacob and Chang, Ming-Wei and Lee, Kenton and Toutanova, Kristina},
  booktitle={NAACL HLT},
  year={2019}
}

@inproceedings{radford2021learning,
  title={Learning Transferable Visual Models From Natural Language Supervision},
  author={Radford, Alec and Kim, Jong Wook and Hallacy, Chris and others},
  booktitle={ICML},
  year={2021}
}

@article{baltrusaitis2018multimodal,
  title={Multimodal Machine Learning: A Survey and Taxonomy},
  author={Baltrušaitis, Tadas and Ahuja, Chaitanya and Morency, Louis-Philippe},
  journal={IEEE Transactions on Pattern Analysis and Machine Intelligence},
  volume={41},
  pages={423–443},
  year={2018}
}

@inproceedings{kiela2020hateful,
  title={The Hateful Memes Challenge: Detecting Hate Speech in Multimodal Memes},
  author={Kiela, Douwe and Firooz, Hamed and Mohan, Aravind and Goswami, Vedanuj and Singh, Amanpreet and Fitzpatrick, Dave and Turian, Joseph},
  booktitle={Advances in Neural Information Processing Systems (NeurIPS)},
  year={2020}
}

@inproceedings{gomez2020multimodal,
  title={Exploring Hate Speech Detection in Multimodal Publications},
  author={Gomez, Raul and others},
  booktitle={IEEE Winter Conference on Applications of Computer Vision (WACV)},
  year={2020}
}

@article{arya2024clip,
  title={Multimodal Hate Speech Detection in Memes Using Contrastive Language-Image Pre-Training},
  author={Arya, A. and others},
  journal={IEEE Access},
  year={2024}
}

@article{zhu2024zeroshot,
  title={Multimodal Zero-Shot Hateful Meme Detection},
  author={Zhu, Y. and others},
  journal={WebSci '22: Proceedings of the 14th ACM Web Science Conference 2022},
  year={2022}
}

@inproceedings{yamagishi2024simple,
  title={Simpler Prompts, Better Results: Enhancing Zero-Shot Detection with a Large Multimodal Model},
  author={Yamagishi, J.},
  booktitle={ACL Workshop Proceedings},
  year={2024}
}

@article{kucerova2020crime,
  title={Machine Learning for Crime Report Classification},
  author={Ku{\v{c}}erov{\'a}, Z. and others},
  journal={IEEE Access},
  year={2020}
}

@article{karystianis2018ie,
  title={Automated Information Extraction from Police Reports},
  author={Karystianis, G. and others},
  journal={Journal of Biomedical Informatics},
  year={2018}
}

@article{du2020forensics,
  title={Deep Learning for Digital Forensics: A Survey},
  author={Du, X. and others},
  journal={Forensic Science International: Digital Investigation},
  year={2020}
}
\end{document}